\documentclass[journal]{IEEEtai}

\usepackage[colorlinks,urlcolor=blue,linkcolor=blue,citecolor=blue]{hyperref}

\usepackage{color,array}

\usepackage{graphicx}
\usepackage{algorithm}
\usepackage{algorithmic}
\usepackage{amsmath}
\usepackage{enumerate}
\usepackage{bbold}

\usepackage{subcaption}
\usepackage{arydshln}
\usepackage{cite}

\begin{document}

\title{GradCFA: A Hybrid Gradient-Based Counterfactual and Feature Attribution Explanation Algorithm for Local Interpretation of Neural Networks}

\author{Jacob Sanderson, Hua Mao, and Wai Lok Woo~\IEEEmembership{Senior Member,~IEEE,}
\thanks{© 2025 IEEE. Personal use of this material is permitted. Permission from IEEE must be obtained for all other uses, in any current or future media, including reprinting/republishing this material for advertising or promotional purposes, creating new collective works, for resale or redistribution to servers or lists, or reuse of any copyrighted component of this work in other works.}
\thanks{This is the accepted version of the paper published in IEEE Transactions on Artificial Intelligence. The final version is available at DOI: \url{http://doi.org/10.1109/TAI.2025.3552057}.}
\thanks{This work was partly funded through the UK government’s Flood and Coastal Resilience Innovation Programme (FCRIP). FCRIP is funded by DEFRA and managed by the Environment Agency.}
\thanks{J. Sanderson, H. Mao, and W.L. Woo are with Department of Computer and Information Sciences, Northumbria University, Newcastle, UK, NE1 8ST\\ (e-mail: jacob.sanderson@northumbria.ac.uk; hua.mao@northumbria.ac.uk; wailok.woo@northumbria.ac.uk).}}

\markboth{DOI: \url{http://doi.org/10.1109/TAI.2025.3552057}}
{J. Sanderson \MakeLowercase{\textit{et al.}}: GradCFA: Hybrid Gradient-Based Counterfactual and Feature Attribution Explanations }

\maketitle

\begin{abstract}
Explainable Artificial Intelligence (XAI) is increasingly essential as AI systems are deployed in critical fields such as healthcare and finance, offering transparency into AI-driven decisions. Two major XAI paradigms, counterfactual explanations (CFX) and feature attribution (FA), serve distinct roles in model interpretability. This study introduces GradCFA, a hybrid framework combining CFX and FA to improve interpretability by explicitly optimizing feasibility, plausibility, and diversity—key qualities often unbalanced in existing methods. Unlike most CFX research focused on binary classification, GradCFA extends to multi-class scenarios, supporting a wider range of applications. We evaluate GradCFA’s validity, proximity, sparsity, plausibility, and diversity against state-of-the-art methods, including Wachter, DiCE, CARE for CFX, and SHAP for FA. Results show GradCFA effectively generates feasible, plausible, and diverse counterfactuals while offering valuable FA insights. By identifying influential features and validating their impact, GradCFA advances AI interpretability. The code for implementation of this work can be found at: \url{https://github.com/jacob-ws/GradCFs}
\end{abstract}

\begin{IEEEImpStatement}
GradCFA offers a robust, locally explainable framework for neural networks by combining counterfactual and feature attribution (FA) methods, optimized for feasibility, plausibility, and diversity. This hybrid approach enhances applicability across sectors such as economics, healthcare, law, and environmental management. By integrating FA, GradCFA provides deeper insights into model behavior, identifying the influence of each feature in counterfactual generation. Representing a technological leap in AI interpretability, GradCFA helps meet legal standards for transparency and addresses the social demand for ethical, accountable AI in high-stakes settings. Its extension to multi-class problems further supports broad, real-world applicability.
\end{IEEEImpStatement}


\section{Introduction}

\IEEEPARstart{A}{s} artificial intelligence (AI) increasingly permeates fields like healthcare \cite{bharati2023review}, cancer detection \cite{shin2023deep}, credit scoring \cite{el2023feature}, loan prediction \cite{zhu2023explainable}, legal decision-making \cite{collenette2023explainable}, and disaster relief \cite{sanderson2023xfimnet}, the need for transparency has become paramount. Explainable AI (XAI) therefore aims to make AI models more interpretable and trusted in these critical applications \cite{saranya2023systematic}.

Feature attribution (FA) is a widely used XAI approach that scores the importance of input features, clarifying model decisions and supporting real-world deployment. Counterfactual explanations (CFX), another XAI technique, explain decisions by identifying input changes that could alter outcomes. This enables users to see cause-effect relations in AI predictions, a high level of interpretability in Pearl’s hierarchy \cite{pearl2018theoretical}, providing actionable information for recourse \cite{jiang2024robust}.

CFX fosters user trust by illustrating how input changes can impact outputs, revealing the model's internal logic in an intuitive way. This capability is increasingly seen as critical in building trustworthy AI systems \cite{del2024generating}. For practical use, CFX must generate feasible and realistic counterfactuals. The CFX community identifies several attributes of effective counterfactuals: validity (whether the counterfactual instance is correctly classified), proximity (closeness to the original instance), sparsity (number of changed features), plausibility (alignment with data distribution), and diversity (range among counterfactuals) \cite{guidotti2024counterfactual}.

Most CFX methods \cite{guidotti2024counterfactual,verma2024counterfactual,jiang2024robust} optimize only a subset of these qualities, often relying on user constraints for feasibility and plausibility, and genetic algorithms for diversity. However, user reliance can limit diversity, while gradient-descent, another optimization method, often restricts exploration and can reach only local optima. Although genetic algorithms can enhance exploration, they may produce less precise results and yield overly numerous solutions, which complicates interpretability.

Despite complementary strengths, FA and CFX have limitations when applied individually. While CFX supports actionable insights, it may be limited in interpreting model behavior, especially when counterfactual sets are large. Here, FA can provide more general interpretability, however, these approaches are limited in their actionable specificity. Existing FA methods like SHAP can be used alongside CFX, but they operate independently, which complicates the implementation by requiring separate integration of multiple post-hoc explainers.

CFX research is further limited, as existing studies are restricted to binary classification \cite{guidotti2024counterfactual,verma2024counterfactual,jiang2024robust,prado2024survey}, making multi-class CFX underexplored, though essential, as many real-world applications involve complex multi-class decision boundaries. In addition, a critical, yet frequently overlooked aspect of CFX research is the explainability of the counterfactual generation process itself. 

In this study, we introduce GradCFA, a hybrid CFX and FA algorithm, that generates counterfactuals optimized for proximity, sparsity, plausibility, and diversity, while providing FA scores to explain model behavior. This integration eliminates the complexity of implementing CFX and FA independently, providing a streamlined solution that supports practical deployment. Furthermore, the FA scores are computed dynamically during counterfactual generation, offering insights into the feature-specific contributions at every step of the optimization process. This ensures transparency in how counterfactuals are derived, enhancing user trust in both the outcomes and the methodology. GradCFA further extends the applicability of CFX to multi-class problems, overcoming the common restriction to binary classification and enabling robust explanations for more complex decision boundaries. These insights are essential in high-stakes contexts, where understanding both counterfactual recommendations and underlying model reasoning supports trust and practical application. GradCFA contributes to the XAI literature by: 
\begin{enumerate}[(i)] 
\item Optimizing validity, proximity, sparsity, plausibility, and diversity, incorporating perturbations to escape local optima and allowing user control over the optimization process. 
\item Providing FA scores by computing gradients with respect to input features, allowing both interpretability and actionable insights, while shedding light on the counterfactual generation process. 
\item Extending CFX to multi-class problems, enhancing trustworthiness and interpretability across a wider range of real-world applications. 
\end{enumerate}

This paper is organized as follows: Section 2 reviews existing work in CFX and FA. Section 3 outlines our methodology, Section 4 presents experimental results, Section 5 presents the discussion, and Section 6 concludes with future directions.

\section{Related Work}
The concept of CFX was first introduced by Wachter et al. \cite{wachter2018counterfactual}, defining a CFX as the closest instance with a different classification from the original. Their method generates counterfactuals close to the original instance but lacks some qualities desirable for robust CFX. DiCE \cite{Mothilal_2020}, a widely used extension, addresses this by enhancing diversity through a loss function term that maximizes pairwise distances between instances, giving users multiple intervention options. While DiCE introduces user constraints for feasibility, it does not account for sparsity or plausibility.

Subsequent approaches build on these limitations, for instance DECE \cite{cheng2021dece} enhances sparsity by restricting changes to the top $k$ influential features, improving feasibility but not plausibility. In \cite{tsiourvas2024manifold} plausibility is considered by optimizing the local outlier factor to ensure counterfactual realism, though this method lacks the diversity and feasibility of prior work.

Genetic algorithms have also been explored for CFX, as in \cite{schleich2021geco}, adapting Wachter's approach to optimize for diverse solutions across a larger search space. While effective in escaping local optima and generating diversity, genetic algorithms tend to be less efficient than gradient-descent-based methods. In \cite{rasouli2021analyzing}, counterfactuals are used to analyze model robustness to perturbation, focusing on proximity and classification constraints. Other multi-objective CFX methods include NSGA-II \cite{dandl2020multi} and CARE \cite{rasouli2024care}, which optimize validity, soundness, and actionability alongside proximity and sparsity. CARE emphasizes coherency between changed and unchanged features, providing user-aligned counterfactuals with a focus on minimizing feature adjustments.

Most existing CFX methods assume a binary classifier, with limited exploration of CFX for multi-class problems. Multi-class counterfactuals in \cite{multiclass} rely on support vector data descriptions (SVDD), which have limitations with complex, high-dimensional data. Similarly, in \cite{electronics12102215} fuzzy decision trees are applied for rule-based multi-class explanations, offering interpretability but limited handling of complex data. Despite advances, there remains a need for methods that directly optimize diversity, plausibility, and feasibility, especially in multi-class contexts.

In FA, popular methods span model-agnostic and model-specific techniques. LIME \cite{ribeiro2016why} uses a surrogate model for local approximations, while Anchors \cite{ribeiro2018anchors} identifies feature conditions for high-confidence predictions. SHAP \cite{lundberg2017a} applies game-theoretic Shapley values for consistent attribution, though it is computationally intensive. Model-specific approaches include Layerwise Relevance Propagation (LRP) \cite{bach2015pixel}, which traces relevance scores, and Grad-CAM \cite{selvaraju2017grad}, which uses heatmaps to show image regions relevant to CNN predictions. Each of these methods has unique strengths and trade-offs, such as LRP’s layer-wise detail and Grad-CAM’s intuitive visualization.

\section{Methodology}

\subsection{Validity}
The validity of a counterfactual refers to whether it is correctly classified as the desired class $y'$. In binary classification, this means the class changes, while in multi-class classification, it changes to a specified value. This can be ensured by using a standard loss function, minimizing the distance between the predicted probability and the target class. GradCFA offers a choice between two loss functions for binary classification problems, hinge loss, and binary cross entropy (BCE) loss. Hinge loss aims to maximise the margin between the two classes, ensuring a clearer boundary between the query instance and counterfactual instances, as follows:

\begin{equation}
    L_{val}^{hinge}=max(0,1-y\cdot f(x'))
\end{equation}

\noindent where $y \in \{-1, +1\}$ is the target outcome, with $-1$ indicating a negative class, and $+1$ indicating positive, and $f(x')$ is the predicted outcome of a counterfactual instance.

BCE loss, on the other hand, quantifies the difference between the probability output by the model with the original class, as follows:
\vspace{-0.5pt}
\begin{equation}
    L_{val}^{BCE} = -\frac{1}{n}\sum^n_{i=1}[y'_{i}log(p_{i})+(1-y'_i)log(1-p_i)]
\end{equation}

\noindent where $n$ is the number of instances in the counterfactual set, $y \in \{0,1\}$ is the target class, with $0$ representing the negative class, and $1$ the positive class, $p$ is the probability of an instance belonging to the positive class, and by extension $1-p$ is the probability of an instance belonging to the negative class. This option is better suited where the goal is to generate counterfactuals that closely resemble the desired class, probabilistically.

\subsection{Feasibility}
The feasibility of a counterfactual refers to how easily a suggested modification could be implemented by a user. To encourage feasible results, we include two terms in the loss function: proximity and sparsity. Proximity ensures that the changes to feature values are small, while sparsity ensures that few features are changed.

\subsubsection{Proximity}
To determine the proximity of the CFX we follow Wachter et al. \cite{wachter2018counterfactual}, and measure the average feature-wise absolute difference between the original instance $x$ and the counterfactual instance $x'$. To ensure the features are within the same scale, each feature is normalized by dividing by its median absolute deviation (MAD) value. The proximity metric is computed as follows:
\vspace{-0.5pt}
\begin{equation}
    L_{prox} = \frac{1}{nm}\sum^n_{i=1}\sum^m_{j=1}\frac{|x'_{i,j} - x_j|}{MAD_j}  
\end{equation}

\noindent where $m$ is the number of features in each instance. 

\subsubsection{Sparsity}
To measure the sparsity of the counterfactuals, we use a boolean operator to determine whether the feature has changed from $x$ to $x'$, meaning the change is greater than $\epsilon$, as defined by the user. We then sum the differences, giving the total number of features changed, then take the average of this value across the counterfactual set, as follows:
\vspace{-0.5pt}
\begin{equation}
    L_{spars} = \frac{1}{nm}\sum^n_{i=1}\sum^m_{j=1}\mathbb{1}(|x'_{ij} - x_j| \geq \epsilon)
\end{equation}

\subsection{Plausibility}
We base our plausibility measurement on the fourth objective proposed by Dandl et al. \cite{dandl2020multi}, which measures the distance between $x'$ with the nearest $k$ instances in the observed data $X$ as an approximation of how likely $x'$ is to be in $X$. We scale this value by the minimum and maximum distances to normalize the loss value between 0 and 1, as follows:
\vspace{-0.5pt}
\begin{equation}\label{plaus}
    L_{plaus} = \frac{1}{n}\sum^n_{i=1}\left(\frac{1}{k}\sum^k_{j=1}\frac{|x'_i-x_j|-d_{min}}{d_{max}-d_{min}+\epsilon}\right)
\end{equation}

\noindent where $d_{min}$ and $d_{max}$ are the minimum and maximum distances among the $k$ nearest instances, respectively. 

\subsubsection{User Constraints}
The plausibility metric given in equation \ref{plaus} provides an approximation of how realistic $x'$ may be. In real-world applications, however, there may be specific feature changes that are not possible. In order to ensure real-world plausibility, we incorporate a range of constraints that enable the user to restrict these changes as necessary. Following DiCE \cite{Mothilal_2020}, we enable users to specify features to vary, as well as acceptable ranges for feature values. By default, all features are permitted to vary, otherwise users can provide a list of features. This is then used to mask the counterfactual instances, such that if a feature is not included in the list its value is replaced with the value from the query instance. For acceptable feature ranges, the default is to bound the features by the minimum and maximum values from the observed dataset, but users can specify a dictionary with the feature name as key, and the acceptable range as the value.

It is beneficial to constrain the features to vary, if the number of features that can be varied is small, but if there are only a few features that must remain fixed then this can be a cumbersome option. As such we propose an additional constraint of features to fix, where users specify a list of features to remain fixed, rather than varied. Finally, there are occasions where a feature may be able to change, but only in one direction. For example, it is plausible for the age of an individual to increase, however it cannot decrease. As such we incorporate an additional constraint on the direction of the feature change, where users specify a dictionary with the feature name as key, and either `increase' or `decrease' as the value.

\subsection{Diversity}
A diverse counterfactual set ensures that a broad range of options are presented to the user, increasing the likelihood of a suitable result being given. This also provides greater insight to approximate the decision boundary of the model. To measure the diversity of the counterfactual set, we follow the diversity metric proposed in \cite{Mothilal_2020}, measuring the pairwise distance between each pair of instances in the set. We construct a matrix of these pairwise distances, then find the determinantal point process (DPP) of this matrix, as follows:
\vspace{-0.5pt}
\begin{equation}
    L_{div} = dpp\left(\frac{1}{1+\sum^m_{l=1}|x'_{il}-x'_{jl}|}\right)
\end{equation}

Unlike the other loss terms (e.g., proximity, sparsity), which are minimized to ensure close, sparse, and plausible counterfactuals, \( L_{div} \) is designed to be maximized. This is because higher values of \( L_{div} \) indicate a larger diversity within the counterfactual set, reflecting greater pairwise differences among instances. By maximizing \( L_{div} \), we encourage counterfactuals that offer distinct alternatives, rather than closely clustered solutions.

\subsection{Optimization}
To optimize the counterfactual set, we construct a loss function from the above defined terms, given in Equation \ref{loss}. Our loss function is a weighted sum of the terms, where users can provide a weight for the proximity, sparsity, plausibility, and diversity terms, enabling them to customize the results to their preferences. We aim to maximise the diversity, so we invert it by taking $1-L_{div}$, such that as this term is minimized, the diversity itself is maximised. The total loss function is computed as follows:
\vspace{-0.5pt}
\begin{equation}\label{loss}
\begin{split}
    L = L_{val}(f(x'),y') + \lambda_{prox} \cdot L_{prox}(X',x) + \\ \lambda_{spars} \cdot L_{spars}(X',x) + 
    \lambda_{plaus} \cdot L_{plaus}(X',X) + \\  \lambda_{div} \cdot (1 - L_{div}(X')) + L_{cat}(X')
\end{split}
\end{equation}

The term $L_{cat}$ is a regularization term for categorical features, to ensure that when one hot encoded, each component of the feature sums to 1. This is enforced through a linear equality constraint, which iterates over each feature and the squared deviation of the sum of the probabilities is computed for each of the categories as follows:
\vspace{-0.5pt}
\begin{equation}
    L_{cat} = \sum_{v\in cat}\sum^n_{i=1}\left(\left(\sum^{v_{last}}_{j=v_{first}}x'_{ij}\right)-1\right)^2
\end{equation}

\noindent where $v \in cat$ denotes the one hot encoded indices of each categorical feature, with $v_{first}$ representing the first index, and $v_{last}$ representing the last index for each categorical feature.

For each of the terms for the characteristics $L_{char} = {L_{prox}, L_{spars}, L_{plaus}}$, if the loss value does not meet a user defined threshold $\tau_{char} = {\tau_{prox}, \tau_{spars}, \tau_{plaus}}$ we add an additional penalty, scaled by user-defined scale factor $\gamma$, to more aggressively optimize that term as follows:
\vspace{-0.5pt}
\begin{equation}
  L_{char}=\begin{cases}
    L_{char}, &  \text{if $L_{char}\leq \tau_{char}$.}\\
    L_{char} \left(1 + \gamma_{char} \right), &  \text{otherwise}.
  \end{cases}
\end{equation}

Since we are maximising diversity $L_{div}$, for this term we apply the penalty if the loss value exceeds the user defined threshold $\tau_{div}$, with the penalty being a subtraction rather than addition, computed as follows:
\vspace{-0.5pt}
\begin{equation}
  L_{div}=\begin{cases}
    L_{div}, &  \text{if $L_{div}\geq \tau_{div}$.}\\
    L_{div} \left(1 - \gamma_{char} \right), &  \text{otherwise}.
  \end{cases}
\end{equation}

We minimize our loss function using the Adam optimizer \cite{kingma2014adam}, which adapts learning rates based on the first and second moments of the gradients, ensuring efficient and robust convergence. A key challenge with gradient-based optimizers is their tendency to get stuck in local optima. To address this, we implement an optimization strategy where, after convergence, we check if the overall loss meets a user-defined threshold $\tau_{loss}$. If the loss exceeds this threshold, indicating suboptimal results, we perturb the feature values to escape the local optima and re-run the optimization. This process repeats until the loss is acceptable or a maximum number of perturbation attempts is reached. The perturbation is achieved by introducing random noise, sampled from a normal distribution, to each feature that differs from the original instance by more than $\epsilon$. We select these features for perturbation such that we do not compromise the sparsity of the returned counterfactual set. This random noise is multiplied by a scale factor $\gamma_{pert}$, defined by the user, and added to the counterfactual feature value as follows:
\vspace{-0.5pt}
\begin{equation}
    x'_i = x'_i + \gamma_{pert} \cdot \mathcal{N}(0,1)
\end{equation}

\noindent where $\mathcal{N}(0,1)$ represents the random noise sampled from a normal distribution. 

\subsection{Handling Multi-Class Problems}
While the majority of work in CFX assumes a binary classifier, we argue that this is not sufficient for practical application, as many crucial real-world problems feature multiple classes, such as income brackets, credit scores, cancer stages and obesity levels. Multi-class classification is a more complex decision-making problem, due to its multiple, often overlapping decision boundaries. Ensuring that GradCFA can handle these problems broadens its real-world utility, offering a flexible option to stakeholders.

The most crucial element to consider when extending from binary to multi-class classification is the validity loss term. This term is fundamental in guiding the correct classification of the generated counterfactual instances. In multi-class problems this differs from binary, as there are multiple class probabilities to consider. For the validity loss term we implement cross entropy loss, as it inherently handles multiple classes by summing the negative log probabilities of the correct class, ensuring that the loss calculation accurately reflects the discrepancy between the predicted and desired classes. Cross entropy loss in multi-class problems is computed as follows:
\vspace{-0.5pt}
\begin{equation}
    L_{val} = -\frac{1}{n}\sum^n_{i=1}\sum^{cl}_{j=1}y'_{ij}log(p_{ij})
\end{equation}

\noindent where $cl$ is the number of classes.

\subsection{Feature Attribution}
\begin{figure*}
    \centering
    \includegraphics[width=0.9\textwidth]{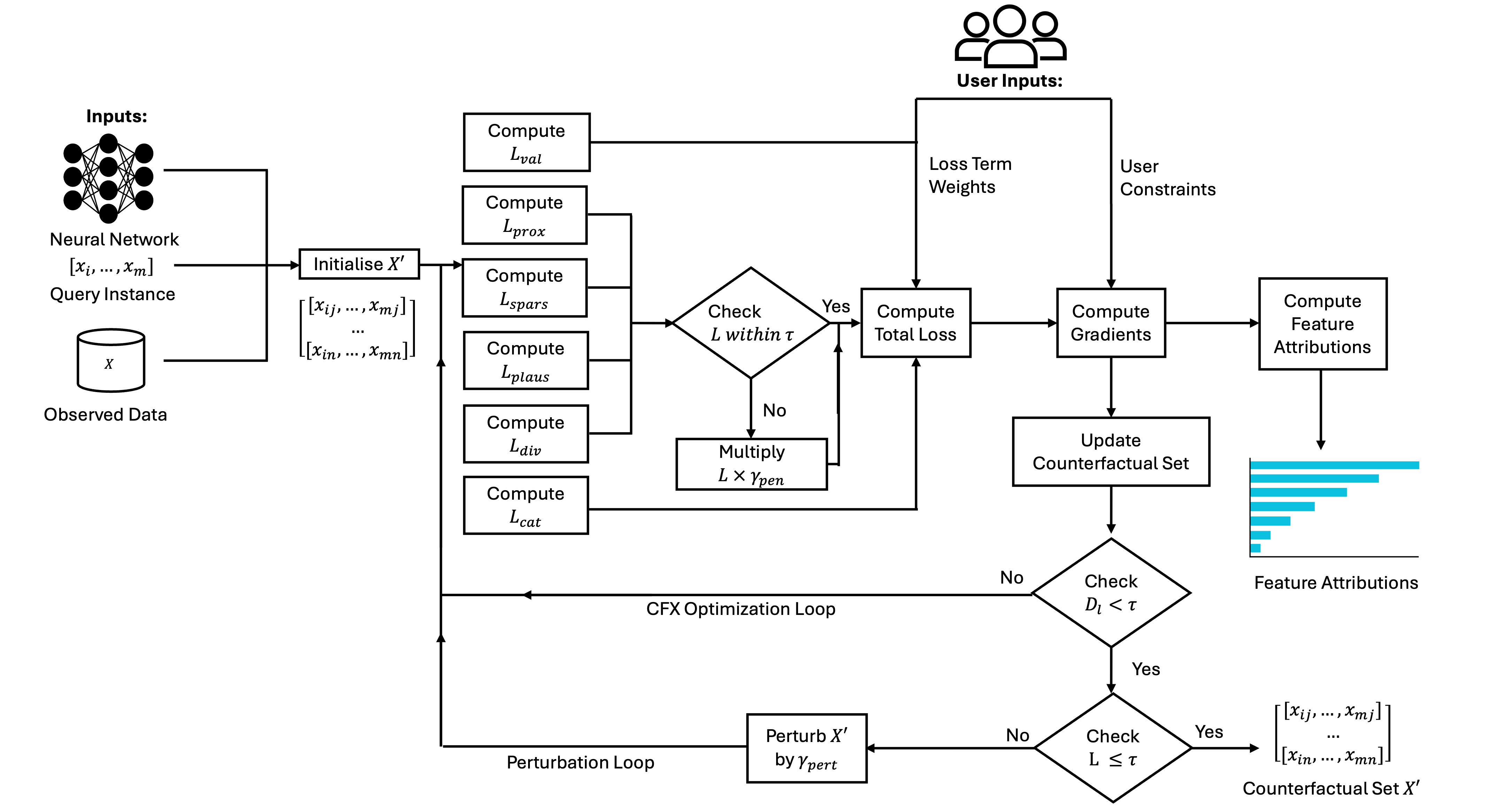}
    \caption{Overview of the GradCFA Pipeline: Beginning with a trained neural network, a query instance, and the observed dataset. The counterfactual set $X'$ is initialized with random values sampled from a normal distribution. Individual loss components are calculated and compared with user-defined thresholds $\tau_{char}$, where each loss component $L$ has its own threshold $\tau_{char}$. For proximity, sparsity, and plausibility $L$ within $\tau$ is true if $L \leq \tau_{char}$, and for diversity $L$ is considered within $\tau$ if $L \geq \tau_{char}$. If thresholds are not met, a penalty is applied. The total loss is computed based on user-defined weighting. User constraints are then applied, gradients are computed to attribute feature importance, and update $X'$. If the final loss after optimization does not meet the threshold, values are perturbed, and optimization is repeated until an acceptable loss is achieved or the maximum number of perturbation attempts is reached, after which the counterfactual set and FAs are returned.}
    \label{fig:GradCFA}
\end{figure*}

To enhance the interpretability of the system, GradCFA also provides FA scores, giving the importance of each feature in the counterfactual generation process. This is achieved by calculating the gradients $\nabla_{x'}f(x')$ which denote the partial derivatives of the predictive model's output $f(x')$ with respect to the input features $x'_i \in x'$. These gradients are computed through backpropagation, and are accumulated across each optimization step to give the cumulative impact of each feature $x'_i$ on the counterfactual prediction. The accumulated gradients are then averaged over the optimization steps, and their magnitude (norm) is used to quantify each feature's importance. The attribution score for each feature $Attr_i$ can therefore be given as follows:
\vspace{-0.5pt}
    \begin{equation}
        \nabla_{ti} = \frac{\partial f(x'_{i,t})}{\partial x'_{i,t}}
    \end{equation}
    \begin{equation}
        Attr_i = \left\|\frac{1}{T}\sum^T_{t=1}\nabla_{ti}\right\|
    \end{equation}

\noindent where $T$ is the total time steps, $x'_t$ is the counterfactual instance at time step $t$. An overview of the GradCFA workflow is given in Algorithm \ref{alg:GradCFA} and Fig. \ref{fig:GradCFA}.

\begin{algorithm}[!htb]
\caption{GradCFA}
\label{alg:GradCFA}
\textbf{Input}: Query Instance $x$, Predictive Model $f(\cdot)$, Hyperparameters $\theta$\\
\textbf{Parameter}: Learning rate $\alpha$, Weights $\lambda$, Thresholds $\tau$, Scale Factors $\gamma$, Maximum Iterations $\mu$, Maximum Perturbation Attempts $\delta$, Number of Counterfactual Instances in the Set $n$\\
\textbf{Output}: Counterfactual Set $X'$, Feature Attributions $Attr$
\begin{algorithmic}[1] 
\STATE Let iteration count $t=0$, perturbation count $p=0$, $X' \leftarrow \mathcal{N}(0,1)^{n \times len(x)}$, $Attr \leftarrow 0$
\WHILE{$t < \delta$ or $D_l \geq \tau_{ld}$}
\STATE Compute loss components $L_{val}$, $L_{di}$, $L_{pl}$, $L_{pr}$, $L_{sp}$, $L_{cat}$.
\FOR{$L_{char}$ in $L_{di}$, $L_{pl}$, $L_{pr}$, $L_{sp}$}
\IF{$L_{char} > \tau_{char}$}
\STATE $L_{char} \leftarrow L_{i}(1-\gamma_{char})$
\ENDIF
\ENDFOR
\STATE Compute total loss $L \leftarrow L_{val} + \lambda_{prox} \cdot L_{prox} + \lambda_{spars} \cdot L_{spars} + \lambda_{plaus} \cdot L_{plaus} + \lambda_{div} \cdot L_{div} + L_{cat}$.
\STATE Compute gradients $\nabla$ of $L$ w.r.t $X'$.
\STATE Compute gradients $\nabla$ of $f(x')$ w.r.t $x'_i$
\STATE Append gradient $\nabla_{x^{'}}f(x^{'})$ to gradients list $Attr$
\STATE Update $X'$ with gradient descent with learning rate $\alpha$.
\STATE Apply user constraints.
\STATE $D_l \leftarrow L_t - L_{t-1}$
\IF {$t \geq \delta$ or $D_l < \tau_{ld}$}
\STATE \textbf{break}
\ENDIF
\STATE $t = t + 1$
\ENDWHILE
\IF {$L > \tau_{pert}$}
\WHILE {$p < \delta$}
\STATE $X' \leftarrow X' + \mathcal{N}(0,1) \cdot \gamma_{pert}$
\STATE $t=0$
\STATE Go to step 2.
\IF {$L \leq \tau_{pert}$}
\STATE \textbf{return} $X'$
\ENDIF
\STATE $p = p + 1$
\ENDWHILE
\ENDIF
\FOR{feature $x_i$, gradient $\nabla_i$ in $Attr$}
\STATE $Attr \leftarrow \left\|\frac{1}{T}\sum^T_{t=1}\nabla_{ti}\right\|$
\ENDFOR
\STATE \textbf{return} $X'$ with lowest $L$
\COMMENT{Return the set of counterfactual instances with the lowest total loss.}
\STATE \textbf{return} $Attr$
\COMMENT{Return the attribution values for each feature.}

\end{algorithmic}
\end{algorithm}

\subsection{Time Complexity}
The GradCFA framework combines CFX and FA, optimizing multiple qualities that contribute to increased computational overhead compared to simpler methods. To evaluate scalability, we analyze the time complexity.

The loss function consists of six components—validity, proximity, sparsity, plausibility, diversity, and categorical regularization—each with distinct computational demands. Validity is computed once per counterfactual, with complexity $O(n)$ where $n$ is the number of counterfactuals. Proximity and sparsity are computed over features, and their complexity is $O(n \cdot m)$. When the feature count $m$ is constant, this simplifies to $O(n)$, however if $m$ grows significantly the complexity scales linearly with $m$. Plausibility requires finding $k$-nearest neighbors, contributing $O(n)$ as $k$ is fixed. The diversity calculation, which involves a matrix determinant, has complexity $O(n^3)$, making it the dominant term as $n$ increases.

During optimization, GradCFA iteratively updates $X'$ to minimize the loss, requiring gradient computation with respect to $X'$. Each gradient step has complexity $O(n)$ with a fixed feature count $m$. Given a maximum iteration count $\mu$, the complexity for optimization becomes $O(\mu \cdot n)$.

The total time complexity of GradCFA is therefore:
\begin{equation}
    O(n^3 + \mu \cdot n)
\end{equation}

The dominant term $O(n^3)$ stems from the diversity component, meaning that scaling primarily depends on $n$. However, $m$ could factor into the time complexity if it grows significantly, affecting both $O(n)$ terms in loss calculation and optimization. For most applications, constants like maximum perturbations $\delta$ remain negligible, but the overall runtime could increase with high feature counts.

Moreover, $n$ should remain relatively low to ensure interpretability, as a large set of counterfactuals could overwhelm users with redundant information. By managing $n$ and setting a reasonable iteration limit $\mu$, GradCFA achieves a balance between computational feasibility and high-quality, interpretable counterfactual explanations.

\subsection{Evaluation}
We evaluate GradCFA on the proximity, sparsity, diversity, and plausibility loss values for the final set of counterfactuals returned. We evaluate the validity of the counterfactuals by measuring the confidence of the classification of the counterfactuals, which we define as the probability assigned by the predictive model to the desired class. For proximity, sparsity, and plausibility, a low value is desirable, while for diversity and validity and high value is preferred. Since the goal is not to excel in any one criteria, but to obtain the best overall balance, we look for the best average of the five main criteria. Since Proximity, Sparsity, and Plausibility are minimized, we use $1-$ each of these values in calculating the average, so a higher average score indicates superior performance. 

\subsubsection{Datasets}
To evaluate our framework we use three datasets, one for binary classification, and two for multi-class, with one having a larger number of features. For binary classification we use the credit approval dataset from the UCI Machine Learning Repository \cite{misc_credit_approval_27}, which contains 690 instances of 14 features, with 4 continuous, and 10 categorical. For multi-class classification we use the obesity levels dataset, also from the UCI Machine Learning Repository \cite{misc_obesity}, which contains 2111 instances of 16 features, 6 of which are continuous, 10 are categorical. This dataset has 7 classes ranging from underweight, to level 3 obesity. We also use a larger dataset from the UCI Machine Learning Repository for the classification of fetal health through cardiotocography data \cite{cardiotocography_193}. This dataset has 2126 instances of 32 features, of which 22 are continuous, and 10 are categorical. This dataset has 3 classes, normal, suspect and pathological. Prior to model training, the continuous features are scaled by the mean and standard deviation, and the categorical features are one hot encoded.

To train the predictive model we split each dataset into training, validation, and testing sets, with a ratio of 60:20:20. GradCFA is evaluated post-hoc on the test set of the pre-trained predictive model. Its performance is assessed by averaging the results of each metric across all test instances, providing a fair representation of its capabilities. The continuous features are scaled using z-score normalization, and categorical features are one hot encoded. To address any missing values in the datasets we drop the corresponding row, as imputation with artificial values may impact the integrity of the counterfactual explanations. 

\subsection{Experimental Setup}
\subsubsection{Predictive Model}
While GradCFA is not a trainable model, it leverages the classification ability of a pretrained predictive model. In our experiments, we use a fully connected neural network with an input layer, followed by 2 hidden layers, the first with 64 neurons, the second with 32, and an output layer. For binary classification a sigmoid activation function is used, while for multi-class we use softmax. We leave the exploration of GradCFA as an explainer for other types of neural network for future work. While the optimization of the predictive model is beyond the scope of this study, it is essential to ensure that the models achieve reasonable accuracy, as this directly impacts the validity and plausibility of the counterfactual results. Table \ref{tab:pred} presents the classification accuracy of the predictive model on the training, validation, and testing sets for all three datasets used in this study. These results indicate that the models are sufficiently accurate to support meaningful counterfactual explanations.

\begin{table}[!ht]
    \centering
    \caption{Classification accuracy of the predictive model on each dataset across training, validation and testing.}
    \begin{tabular}{c|ccc}
        \hline
        Dataset & Training & Validation & Testing  \\
        \hline
         Credit Approval & 0.88 & 0.86 & 0.82 \\
         Obesity & 0.90 & 0.84 & 0.85 \\
         Fetal Health & 0.92 & 0.93 & 0.91 \\
         \hline
    \end{tabular}
    \label{tab:pred}
\end{table}

\subsubsection{Hyperparameter Tuning}
GradCFA has several hyperparameters to tune, we select the values for each hyperparameter through a grid search of different values over the test set, with the following ranges:

\begin{itemize}
    \item $\lambda$: [0.25, 0.5, 0.75, 1.0]
    \item $\gamma_{pen}$: [0.05, 0.1, 0.15, 0.2]
    \item $\gamma_{pert}$: [0.3, 0.5, 0.7, 0.9]
    \item $\tau_{prox}, \tau_{spars}, \tau_{plaus}$: [0.1, 0.2, 0.3, 0.4, 0.5]
    \item $\tau_div$: [0.5, 0.6, 0.7, 0.9]
    \item $\tau_{loss}$: [0.3, 0.4, 0.5, 0.6, 0.7]
\end{itemize}

For $\lambda$ we choose a value of 0.5 for each term, giving the best balance of the proximity, sparsity, plausibility and diversity metrics with the validity of the counterfactuals, while ensuring uniformity for comparison, in particular with previous work. We find a $\gamma_{pen}$ value of 0.1 to provide the best scaling for the additional penalty, striking a balance between being suitably aggressive to optimize the characteristics, while maintaining throughout the process. For the credit approval dataset we find $\tau_{prox}$ and $\tau_{spars}$ values of 0.2, $\tau_{plaus}$ of 0.4, and $\tau_{div}$ of 0.9 to be the most optimal. For the obesity levels problem we find the optimal values to be 0.3 for $\tau_{prox}$ and $\tau_{spars}$, and 0.4 for $\tau_{plaus}$ and $\tau_{div}$. For the fetal health dataset we find $\tau_{prox}$ and $\tau_{spars}$ of 0.4, $\tau_{plaus}$ of 0.5 to provide optimal results. For all datasets, we find a $\gamma_{pert}$ value of 0.5 to provide a suitably large perturbation to escape local optima, while still maintaining the knowledge gained through the previous optimization step. For credit approval, a $\tau_{loss}$ of 0.5 is selected for optimal results, while for obesity a value of 0.4, and for fetal health a slightly higher value of 0.6 is necessary.

\subsubsection{Baseline Models}
In our experiments we compare our work with three benchmark CFX algorithms, including Wachter \cite{wachter2018counterfactual}, DiCE \cite{Mothilal_2020} and CARE \cite{rasouli2024care}. Wachter serves as our baseline due to its seminal status in counterfactual generation, upon which many subsequent algorithms are founded. DiCE is chosen as it remains the state-of-the-art and is widely recognized as the leading algorithm for CFX generation tasks. Lastly, CARE is selected as it represents a recent advancement in CFX, and also provides a view of how the use of a genetic multi-objective algorithm performs in comparison with gradient descent-based approaches. Each algorithm is implemented using Python libraries published by the authors.

\section{Results}

\begin{table*}[!ht]
    \centering
    \caption{Proximity, Sparsity, Plausibility, Diversity and Confidence of GradCFA in Comparison with the results obtained in the experiments conducted in subsections IV A, B, C, E, and F. Proximity, Sparsity and Plausibilty are minimized, while Diversity and Validity are maximized. An overall average of these metrics is given for each experiment.}
    \setlength\tabcolsep{3pt}
    \begin{tabular}{c|c|ccccccc|c|c|ccc|ccccccc}
         \hline
         & & \multicolumn{7}{c|}{Ablation Study Combinations} & \multicolumn{2}{c|}{GradCFA} & \multicolumn{3}{c|}{Previous Work} & \multicolumn{7}{c}{Multi-Class} \\
         \hline
         & GradCFA & 1 & 2 & 3 & 4 & 5 & 6 & 7 & W/o Pert. & W/o Pen. & Wachter & DiCE & CARE & 0 & 1 & 2 & 3 & 4 & 5 & 6\\
         \hline
         \multicolumn{20}{c}{Credit Approval Dataset for Binary Classification}\\
         \hline
         Prox. & 0.15 & 0.14 & 0.12 & 0.19 & 0.16 & 0.16 & 0.13 & 0.23 & 0.23 & 0.21 & 0.03 & 0.12 & 0.18 & - & - & - & - & - & - & - \\
         Spars & 0.18 & 0.19 & 0.16 & 0.21 & 0.20 & 0.20 & 0.17 & 0.25 & 0.27 & 0.24 & 0.10 & 0.28 & 0.56 & - & - & - & - & - & - & - \\
         Plaus & 0.37 & 0.68 & 0.53 & 0.61 & 0.46 & 0.45 & 0.54 & 0.42 & 0.51 & 0.45 & 0.56 & 0.51 & 0.47 & - & - & - & - & - & - & - \\
         \hdashline
         Div & 0.96 & 0.86 & 0.72 & 0.82 & 0.92 & 0.80 & 0.88 & 0.94 & 0.91 & 0.92 & 0.35 & 0.84 & 0.68 & - & - & - & - & - & - & -\\
         Conf. & 0.78 & 0.76 & 0.74 & 0.76 & 0.71 & 0.78 & 0.73 & 0.72 & 0.75 & 0.70 & 0.56 & 0.64 & 0.66 & - & - & - & - & - & - & - \\
         \hdashline
         Avg. & 0.81 & 0.76 & 0.73 & 0.71 & 0.76 & 0.75 & 0.75 & 0.75 & 0.73 & 0.74 & 0.64 & 0.71 & 0.63 & - & - & - & - & - & - & - \\
         \hline
         \multicolumn{20}{c}{Obesity Levels Dataset for Multi-Class Classification}\\
         \hline
         Prox. & 0.21 & 0.06 & 0.24 & 0.26 & 0.17 & 0.24 & 0.16 & 0.26 & 0.35 & 0.37 & - & - & - & 0.21 & - & 0.21 & 0.22 & 0.26 & 0.29 & 0.35\\
         Spars & 0.23 & 0.25 & 0.25 & 0.32 & 0.29 & 0.25 & 0.36 & 0.20 & 0.35 & 0.40 & - & - & - & 0.25 & - & 0.23 & 0.27 & 0.23 & 0.26 & 0.31\\
         Plaus & 0.34 & 0.50 & 0.70 & 0.52 & 0.50 & 0.53 & 0.48 & 0.47 & 0.45 & 0.51 & - & - & - & 0.39 & - & 0.34 & 0.52 & 0.59 & 0.61 & 0.67\\
         \hdashline
         Div & 0.92 & 0.04 & 0.11 & 0.79 & 0.84 & 0.64 & 0.42 & 0.82 & 0.80 & 0.83 & - & - & - & 0.90 & - & 0.92 & 0.87 & 0.84 & 0.85 & 0.83\\
         Conf. & 0.76 & 0.68 & 0.79 & 0.74 & 0.71 & 0.78 & 0.68 & 0.76 & 0.73 & 0.69 & - & - & - & 0.71 & - & 0.76 & 0.73 & 0.70 & 0.67 & 0.64\\
         \hdashline
         Avg. & 0.78 & 0.58 & 0.54 & 0.69 & 0.72 & 0.68 & 0.62 & 0.73 & 0.68 & 0.65 & - & - & - & 0.75 & - & 0.78 & 0.72 & 0.69 & 0.67 & 0.63\\
         \hline
         \multicolumn{20}{c}{Fetal Health Dataset for Multi-Class Classification}\\
         \hline
         Prox. & 0.34 & 0.29 & 0.33 & 0.47 & 0.46 & 0.35 & 0.29 & 0.46 & 0.40 & 0.45 & - & - & - & - & 0.34 & 0.36 & - & - & - & -\\
         Spars & 0.35 & 0.51 & 0.31 & 0.52 & 0.51 & 0.40 & 0.33 & 0.52 & 0.51 & 0.49 & - & - & - & - & 0.35 & 0.39 & -& - & - &-\\
         Plaus & 0.42 & 0.56 & 0.51 & 0.42 & 0.49 & 0.48 & 0.63 & 0.45 & 0.54 & 0.47 & - & - & - & - & 0.42 & 0.52 &- &- &- & -\\
         \hdashline
         Div & 0.96 & 0.82 & 0.86 & 0.89 & 0.90 & 0.91 & 0.83 & 0.95 & 0.94 & 0.93 & - & - & - & - & 0.96 & 0.93 & - & - &- & - \\
         Conf. & 0.79 & 0.78 & 0.79 & 0.74 & 0.72 & 0.78 & 0.73 & 0.74 & 0.71 & 0.72 & - & - & - & - & 0.79 & 0.76 & -& -& -&- \\
         \hdashline
         Avg. & 0.73 & 0.65 & 0.70 & 0.64 & 0.63 & 0.69 & 0.66 & 0.65 & 0.64 & 0.65 & - & - & - & - & 0.73 & 0.68 & - & - & - & -\\
         \hline
    \end{tabular}
    \label{tab:results}
\end{table*}

\subsection{Ablation Study}
To demonstrate the improved quality achieved through the inclusion of sparsity, plausibility and diversity, we perform an ablation study. Given the definition of a CFX as the smallest changes to the input features required to change the classification of the model, validity and proximity are integral components of a CFX algorithm, so we include these two components in each combination. Therefore, the following combinations are considered: 1) Validity and Proximity only, 2) Validity, Proximity, and Sparsity, 3) Validity, Proximity, and Plausibility, 4) Validity, Proximity, and Diversity, 5) Validity, Proximity, Sparsity, and Plausibility, 6) Validity, Proximity, Sparsity, and Diversity, 7) Validity, Proximity, Plausibility, and Diversity, and 8) Validity, Proximity, Sparsity, Plausibility, and Diversity (GradCFA).

The results of this ablation study are provided in Table \ref{tab:results}, under the heading `Ablation Study Combinations', with combination 8 under the `GradCFA' heading. This Table shows that for credit approval, combinations 1, 2, and 6 produce the closest and most sparse counterfactuals. However, these combinations suffer in plausibility, diversity, and validity. Combinations 3, 4, and 5 offer moderate proximity and sparsity, with slight improvements in plausibility for combinations 3 and 5, though still weak overall. Combination 7 excels in plausibility and diversity but trades off proximity and sparsity, showing the weakest performance in these areas. Combination 8, the proposed loss function in GradCFA, strikes a balance, achieving moderate proximity and sparsity while excelling in plausibility, diversity, and validity.

For obesity levels, combination 1 again offers the lowest proximity, with combinations 2 and 6 also performing well in this regard but poorly in diversity. Combinations 3, 4, and 5 achieve moderate proximity, but 3 and 5 show low diversity, particularly compared to their performance on the credit approval data. These combinations also have better plausibility but higher sparsity scores. Combination 8, the GradCFA function, balances all metrics, with the best plausibility, strong sparsity and diversity, and moderate proximity. 

In the fetal health dataset, we observe that again, combination 1, 2, and 6 maintain the lowest proximity and sparsity scores, with relatively weak diversity and plausibility. Combinations 3, 4, 5, and 7 all achieve comparable performance across proximity, plausibility, and diversity, however combination 5 achieves a much better sparsity score. As in the other two datasets, GradCFA achieves the best balance across all metrics, with the highest validity and diversity scores, lowest plausibility score, and moderate proximity and sparsity. 

\subsection{Perturbation Analysis}
\begin{figure*}[ht!]
    \centering
    \begin{subfigure}{0.33\textwidth}
    \includegraphics[width=\linewidth]{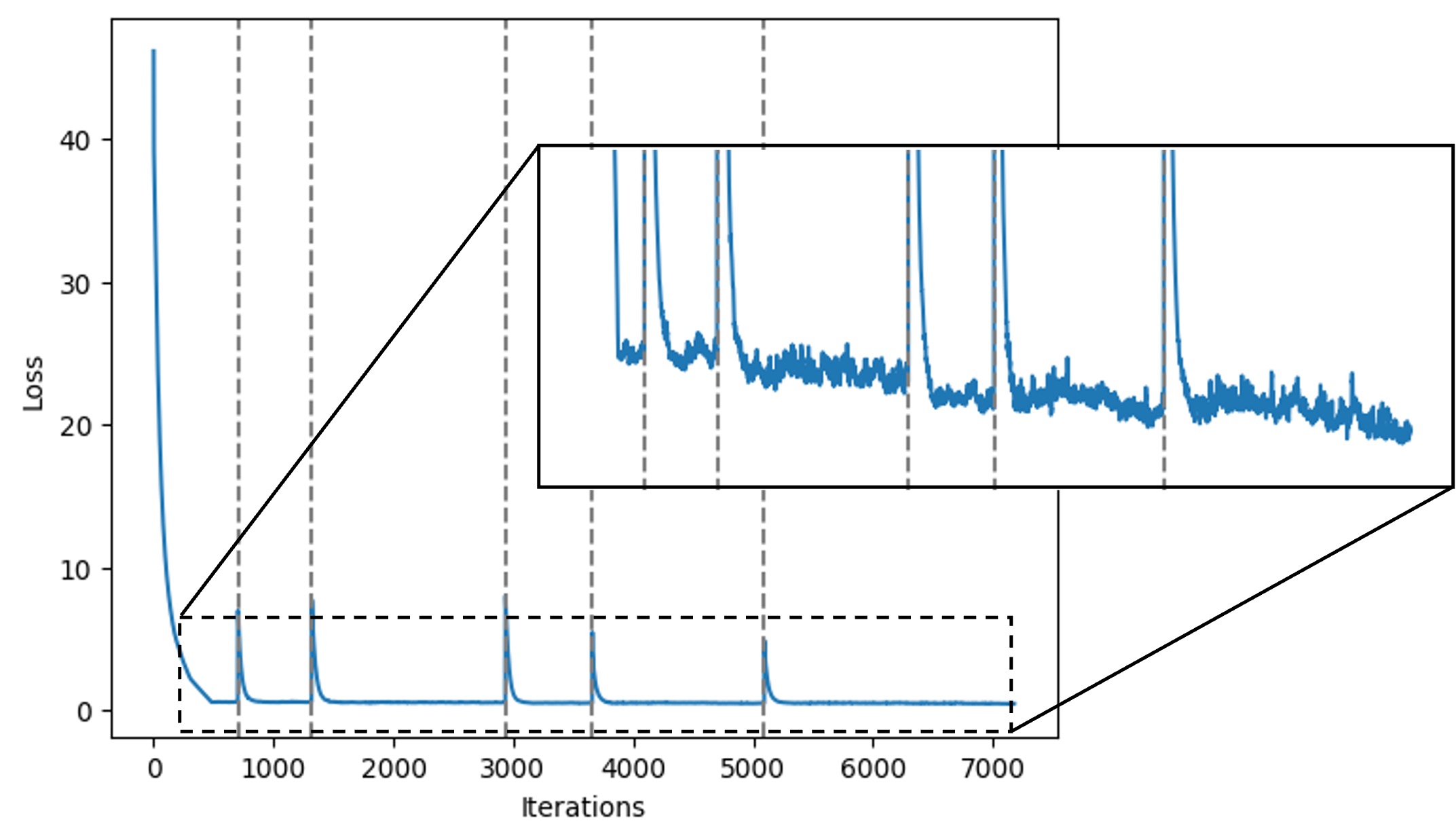}
        \caption{}
    \end{subfigure}
    \begin{subfigure}{0.33\textwidth}
    \includegraphics[width=\linewidth]{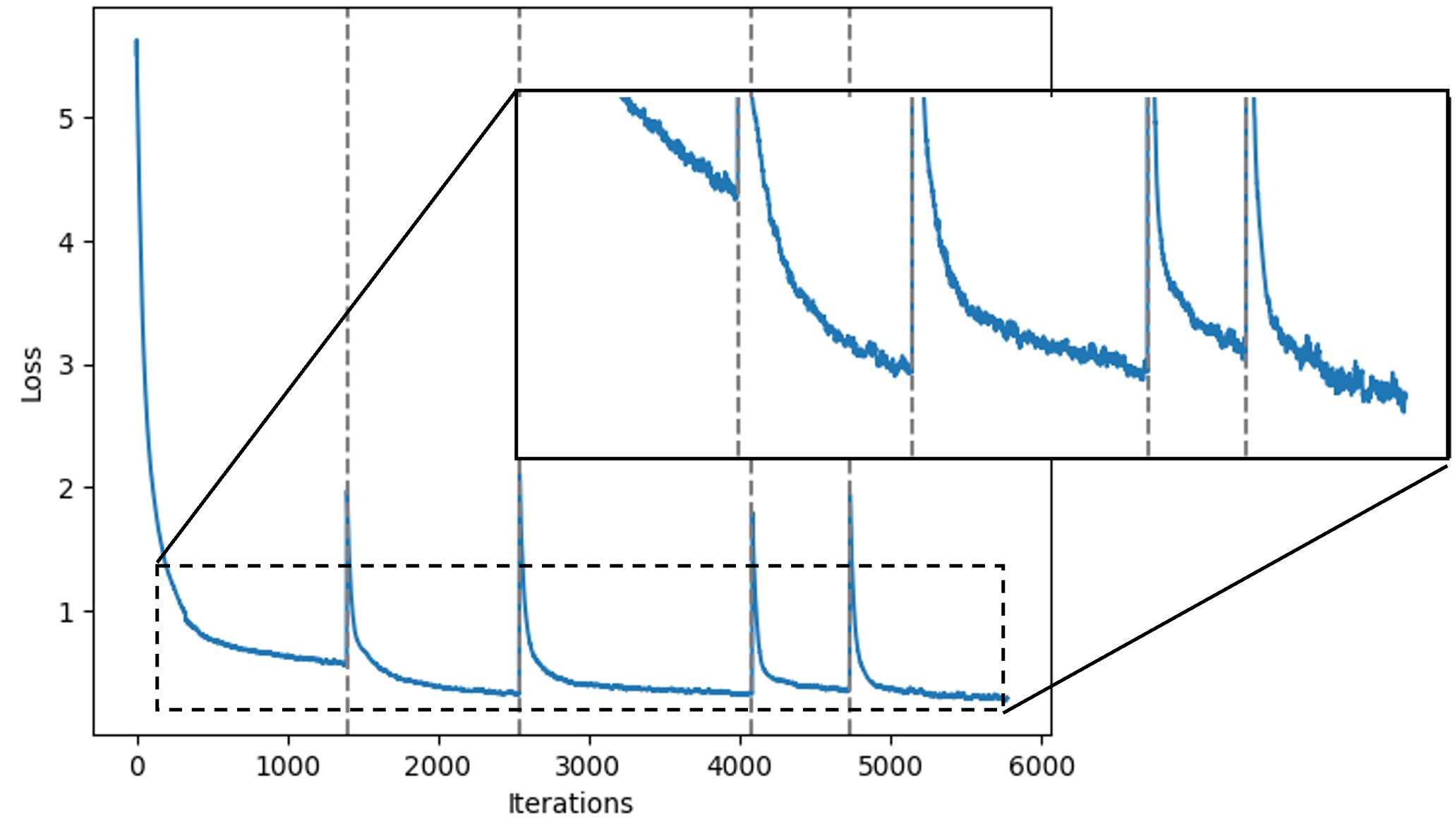}
        \caption{}
    \end{subfigure}
    \begin{subfigure}{0.29\textwidth}
    \includegraphics[width=\linewidth]{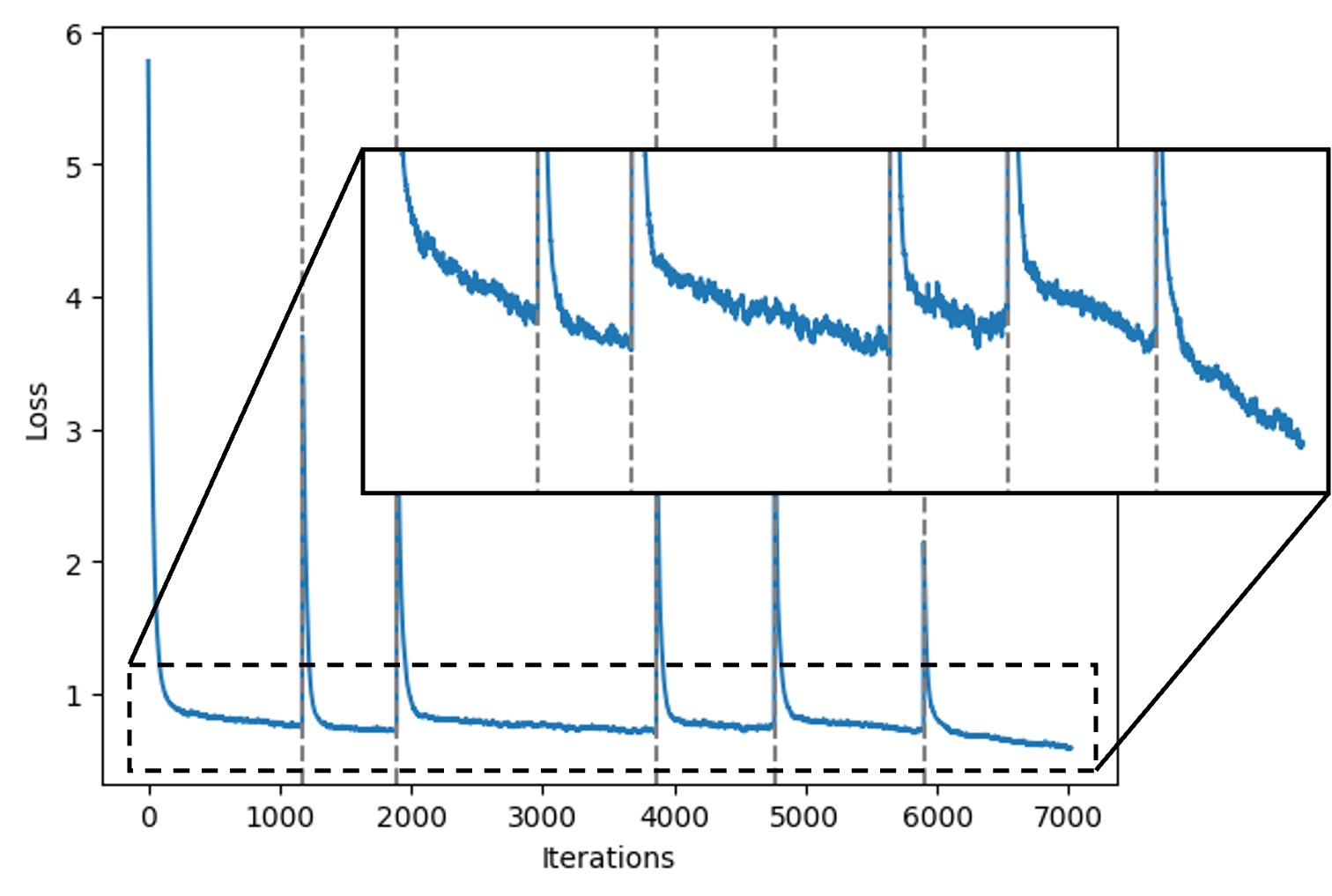}
        \caption{}
    \end{subfigure}
    \caption{Loss curves for (a) credit approval, (b) obesity levels, and (c) fetal health, with perturbation, where the grey dashed lines represent the points of perturbation. The curves demonstrate that after each perturbation the loss is driven further down, indicating the efficacy of the approach in improving the results.}
    \label{fig:lc}
\end{figure*}

\begin{figure*}[ht!]
    \centering
    \begin{subfigure}{0.3\textwidth}
    \includegraphics[width=\linewidth]{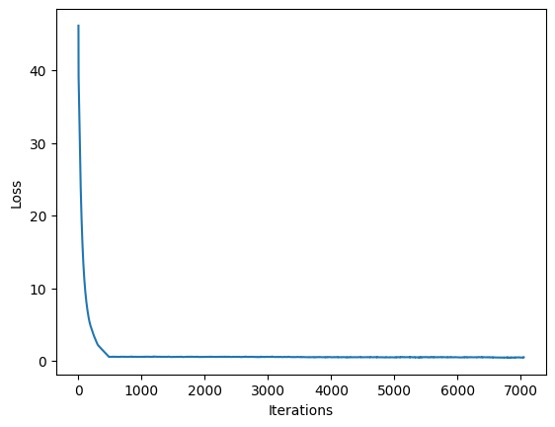}
        \caption{}
    \end{subfigure}
    \begin{subfigure}{0.3\textwidth}
    \includegraphics[width=\linewidth]{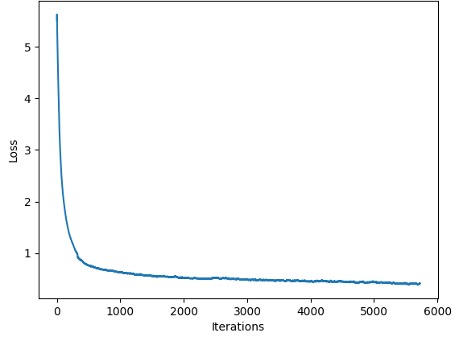}
        \caption{}
    \end{subfigure}
    \begin{subfigure}{0.29\textwidth}
    \includegraphics[width=\linewidth]{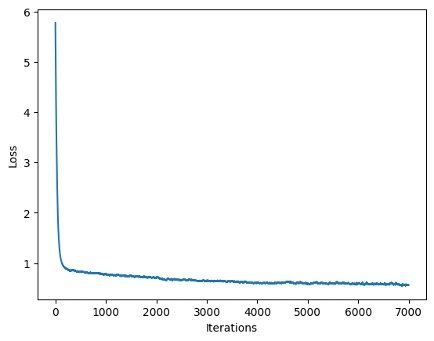}
        \caption{}
    \end{subfigure}
    \caption{Loss curves for (a) credit approval, (b) obesity levels, and (c) fetal health without perturbation.}
    \label{fig:lc2}
\end{figure*}

To assess the effectiveness of perturbation in escaping local optima and improving counterfactual quality, Fig. \ref{fig:lc} shows the loss curve throughout optimization for each dataset. Fig. \ref{fig:lc2} shows the loss curve for the same number of iterations, without perturbation. Additionally, Table \ref{tab:results} shows the results with the perturbation strategy, as `GradCFA' as well as the results without the perturbation strategy as `W/o Pert'. The loss curves show that perturbation enables the algorithm to achieve a lower overall loss, resulting in higher-quality counterfactuals. This confirms the strategy's effectiveness in escaping local optima and achieving more globally optimal results. Across all datasets, the loss value for every metric is improved where perturbation is included in the optimization strategy, further demonstrating the efficacy of this approach.

\subsection{Additional Penalty}
In Table \ref{tab:results} the performance of GradCFA where the additional penalty is included in the terms $L_{prox}, L_{spars}, L_{plaus}$ and $L_{div}$ is presented under `W/o Pen'. Across all datasets, it can be observed that the inclusion of the additional penalty, shown under `GradCFA' is effective in driving a better performance against each metric. Additionally, all of the results fall within the threshold value provided, demonstrating how effectively the performance can be controlled through GradCFA.

\subsection{Example Counterfactuals}
In Tables \ref{tab:cfsc}, \ref{tab:cfso}, and \ref{tab:cfsf}, example counterfactual sets are provided for the credit approval, obesity levels, and fetal health datasets, respectively. For the credit approval dataset, the initial query instance was classified as 0 (rejection), and counterfactuals were generated for class 1 (approval). For the obesity dataset, we used a query classified as 2 (overweight level 1) and aimed to generate counterfactuals for class 1 (normal weight). Finally, for the fetal health dataset, we used a query initially classed as 1 (suspect) and targeted counterfactuals for class 0 (normal).

In the credit approval dataset, certain features such as Age and Income stay relatively similar to the original instance across counterfactuals. Variables like marital status (MS), bank customer status (BC), and citizenship also remain unchanged, suggesting these may have limited influence on credit approval. Gender and ethnicity vary minimally as well, implying that the model is not heavily biased by demographic details in making its decisions. However, Employment Status (ES) consistently shifts from unemployed to employed in counterfactuals, highlighting its significant impact on approval—a pattern that aligns well with real-world expectations. Years employed (YE) tends to increase across the counterfactuals, supporting the idea that extended employment is favorable for credit approval. Conversely, debt levels and previous defaults (PD) rise in each instance, which is counterintuitive as lower debt would generally be preferred. This suggests either an inconsistency in model learning or complex interdependencies among features, where high debt or defaults are acceptable when compensated by other strong indicators.

In the obesity levels dataset, body mass index (BMI) serves as the primary measure, dependent on both height and weight. Here, the generated counterfactuals reflect substantial weight loss for the shift from overweight to normal weight. Interestingly, there is also a minor reduction in height, which is logically inconsistent with the objective of achieving a lower BMI, however, when coupled with the lower weight this change is still reasonable. Most other features remain consistent, except for the frequency of physical activity (FAF), which decreases in each counterfactual. This unexpected outcome may point to feature dependencies within the model, where reduced physical activity is compensated by significant weight loss.

In the fetal health dataset, where the query instance initially indicated a suspect classification, counterfactuals for the normal classification reveal several shifts in clinical measurements. Features related to long term variability (ALTV, MLTV) generally increase, which suggests that the fetal heart rate is adapting dynamically to conditions, reflecting a healthy autonomic nervous system and adequate oxygenation. Conversely, uterine contractions (UC) and deceleration episodes (DL) tend to decrease, aligning with expectations of healthier outcomes. Other features, including accelerations (AC) and mean value (Mean), exhibit variability but follow consistent patterns across counterfactuals, indicating their role in distinguishing suspect from normal fetal health.

\begin{table}[!ht]
    \centering
    \caption{Example Counterfactual Set for Credit Approval.}
    \begin{tabular}{cc|ccccc}
    \hline
         Feature & Query & \multicolumn{5}{c}{Counterfactual Values} \\
         \hline
         Age & 34.14 & 28.55 & 33.44 & 35.34 & 37.26 & 46.75 \\
         Debt & 2.75 & 4.2 & 3.9 & 3.4 & 5.1 & 8.0 \\
         YE & 2.5 & 3.28 & 3.93 & 2.94 & 3.54 & 2.5 \\
         Income & 200.01 & 199.00 & 200.00 & 205.00 & 105.00 & 195.00 \\
         Gender & 1 & 1 & 1 & 1 & 0 & 0 \\
         MS & 1 & 1 & 1 & 1 & 1 & 1 \\
         BC & 1 & 1 & 1 & 1 & 1 & 1 \\
         Industry & 1 & 0 & 1 & 1 & 13 & 1 \\
         Ethnicity & 0 & 4 & 0 & 0 & 0 & 0 \\
         PD & 0 & 1 & 1 & 1 & 1 & 1 \\
         ES & 0 & 1 & 1 & 1 & 1 & 0 \\
         CS & 0 & 2 & 0 & 10 & 0 & 0 \\
         DL & 1 & 1 & 1 & 0 & 1 & 0 \\
         Citizen & 0 & 0 & 0 & 0 & 0 & 0 \\
         Outcome & 0 & 1 & 1 & 1 & 1 & 1 \\
         \hline
    \end{tabular}
    \label{tab:cfsc}
\end{table}

\begin{table}[!ht]
    \centering
    \caption{Example Counterfactual Set for Obesity Levels.}
    \begin{tabular}{cc|ccccc}
    \hline
         Feature & Query & \multicolumn{5}{c}{Counterfactual Values} \\
         \hline
         Age & 19.0 & 20.0 & 19.0 & 21.0 & 23.0 & 22.0 \\
         Height & 1.75 & 1.68 & 1.72 & 1.65 & 1.66 & 1.65 \\
         Weight & 100.0 & 53.4 & 51.9 & 62.3 & 51.6 & 61.8 \\
         FVFC & 2.0 & 1.9 & 1.9 & 1.9 & 2.0 & 2.0 \\
         NCP & 3.0 & 3.0 & 3.0 & 2.9 & 3.0 & 3.0 \\
         CH2O & 2.0 & 1.9 & 2.0 & 2.0 & 2.0 & 2.0 \\
         FAF & 2.0 & 0.6 & 1.2 & 0.8 & 0.4 & 0.5 \\
         TUE & 4.2e-8 & 4.1e-8 & 4.2e-8 & 4.2e-8 & 4.2e-8 & 4.2e-8 \\
         Gender & 1 & 1 & 1 & 1 & 1 & 1 \\
         CALC & 3 & 2 & 3 & 3 & 3 & 2 \\
         FAVC & 1 & 1 & 1 & 1 & 0 & 0 \\
         SCC & 0 & 0 & 0 & 0 & 0 & 0 \\
         SMOKE & 0 & 0 & 0 & 0 & 0 & 0 \\
         FHWO & 1 & 1 & 1 & 1 & 1 & 1 \\
         CAEC & 1 & 1 & 1 & 2 & 1 & 2 \\
         MTRANS & 3 & 3 & 3 & 3 & 3 & 3 \\
         Outcome & 2 & 1 & 1 & 1 & 1 & 1 \\
         \hline
    \end{tabular}
    \label{tab:cfso}
\end{table}

\begin{table}[!ht]
    \centering
    \caption{Example Counterfactual Set for Fetal Health.}
    \setlength\tabcolsep{4pt}
    \begin{tabular}{cc|ccccc}    
    \hline
         Feature & Query & \multicolumn{5}{c}{Counterfactual Values} \\
         \hline
         LBE & 125.0 & 128.9 & 148.0 & 128.3 & 141.5 & 142.2 \\
         LB & 125.0 & 132.8 & 124.9 & 124.8 & 132.1 & 118.2 \\
         AC & 10.0 & 5.1 & 2.8 & 0.1 & 5.5 & 2.2 \\
         FM & -4.6e-11 & -3e-11 & -7.4e-11 & -1.1e-11 & 1.5e-11 & 1.2e-11 \\
         UC & 2.3e-10 & 3e-10 & -1.2e-10 & 4.3e-10 & -4e-10 & -2e-10 \\
         ASTV & 41.0 & 56.9 & 55.1 & 52.6 & 44.2 & 41.0 \\
         MSTV & 1.2 & 0.2 & 0.7 & 0.8 & 1.0 & 1.2 \\
         ALTV & -2.2e-7 & 2.1e-7 & -1.4e-7 & -2.7e-7 & 2.4e-7 & -3.1e-7 \\
         MLTV & 11.4 & 21.1 & 13.2 & 23.5 & 27.6 & 13.6 \\
         DL & -2.5e-9 & -8.9e-9 & -1e-9 & -4.8e-9 & -5.5e-9 & -8.1e-9 \\
         DS & -5.7e-11 & -1.9e-11 & -2.3e-11 & -1.4e-11 & -2.9e-11 & -2.1e-11 \\
         DP & 3.2e-10 & 1.8e-10 &  1.9e-10 & 9.7e-11 & 2.1e-10 & 1.5e-10 \\
         DR & 0.0 & 0.1 & -0.1 & 0.0 & 0.2 & -0.1 \\
         Width & 63.0 & 13.5 & 16.3 & 3.2 & 62.2 & 30.3 \\
         Min & 98.0 & 151.3 & 117.3 & 92.4 & 121.4 & 128.1 \\
         Max & 161.0 & 143.8 & 134.7 & 137.9 & 128.6 & 129.9 \\
         Nmax & 4.0 & 3.7 & 1.1 & 2.9 & 3.8 & 4.3 \\
         Nzeros & -4.1e-9 & -5e-9 & -8.3e-9 & -6.1e-9 & -7.5e-9 & -7.5e-9 \\
         Mode & 138.0 & 119.8 & 113.4 & 125.5 & 126.8 & 131.9 \\
         Mean & 135.0 & 134.6 & 113.1 & 113.6 & 88.4 & 116.4 \\
         Median & 137.0 & 120.7 & 128.9 & 132.2 & 140.9 & 136.9 \\
         Variance & 6.0 & 6.2 & 2.9 & 5.1 & 7.5 & 3.9 \\
         Tendency & 1 & 1 & 0 & 1 & 1 & 0 \\
         A & 0 & 0 & 0 & 0 & 0 & 0 \\
         B & 1 & 1 & 1 & 1 & 1 & 1 \\
         C & 0 & 0 & 0 & 0 & 0 & 0 \\
         D & 1 & 1 & 0 & 1 & 1 & 0 \\
         E & 0 & 0 & 0 & 0 & 0 & 0 \\
         AD & 0 & 0 & 0 & 0 & 0 & 0 \\
         DE & 0 & 0 & 1 & 0 & 0 & 0 \\
         LD & 0 & 0 & 0 & 0 & 0 & 0 \\
         FS & 0 & 0 & 0 & 0 & 0 & 0 \\
         SUSP & 0 & 0 & 0 & 0 & 0 & 0 \\
         Outcome & 1 & 0 & 0 & 0 & 0 & 0 \\
         \hline
    \end{tabular}
    \label{tab:cfsf}
\end{table}

\subsection{Previous Work}
To contextualise the performance of GradCFA, we compare its performance with existing CFX algorithms in Table \ref{tab:results} for the credit approval dataset. As far as the authors are aware, GradCFA is the first CFX algorithm for multi-class problems, therefore we cannot compare with previous work on the obesity levels or fetal health data. In proximity, Wachter achieves the best result, which is expected given that proximity is the main focus of this algorithm. This is followed by DiCE, which focuses on balancing proximity and diversity. CARE and GradCFA both achieve a more modest value on this metric, which can be attributed to the increased complexity in their loss functions. Wachter also achieves the best sparsity score, an intuitive extension of close proximity, as the best proximity score is an unchanged feature. GradCFA follows this, demonstrating the benefit of explicitly optimizing for this where the loss function is more complex. DiCE does not optimize for sparsity, and achieves a slightly weaker performance, demonstrating that there is a likely trade-off in sparsity when optimizing for diversity. CARE performs the weakest on this metric, with more than half of the features changing on average.

GradCFA significantly outperforms the other algorithms on plausibility, demonstrating that the results generated by this algorithm are closer to the distribution of the observed data, so are more likely to exist as real instances. DiCE and Wachter both perform comparably, with DiCE achieving a marginally lower value. CARE does consider the closeness to the observed data in its validity module, which can be attributed to its lower plausibility value. The complexity of CARE, however, does not enable sufficiently efficient optimization of each component for it to achieve comparable performance. GradCFA also achieves the best diversity score, demonstrating its ability to provide a broad range of counterfactual options to the user. Despite DiCE having a greater focus on diversity, with fewer competing objectives to optimize for, it follows GradCFA with a diversity score of 0.12 less. CARE follows DiCE with a much lower diversity score. CARE does not explicitly optimize for diversity, but rather relies on the use of genetic optimization to ensure diversity, which indicates that genetic optimization is not as effective in generating diverse results as a diversity loss function term. Finally, Wachter achieves the poorest diversity score, which is expected, as this algorithm was not intended to provide a diverse set of counterfactuals, but rather the closest possible counterfactual instance. Overall the performance of GradCFA is much more balanced across all four characteristics than any of the previous works, demonstrating its efficacy in optimizing for feasibility, plausibility, and diversity.

\subsection{Multi-Class Analysis}
A key contribution of our work is its extension of CFX to multi-class problems, so here we evaluate the impact of the class of the original instance on the performance against the evaluation metrics. For obesity, we consider the target class to be 1, which is normal weight, and evaluate counterfactuals generated from query instances of each of the other classes in the dataset. For fetal health, we consider the target class to be 0, which is a normal fetus. The results of this evaluation are shown in Table \ref{tab:results} under the heading `Multi-Class'. The results show that as the original class gets further from the target class, the proximity of the results gets further in turn. This is an intuitive finding, as if we consider the real world scenario of obesity levels, an individual who is classified as level 2 obese will need to lose more weight, and make greater lifestyle changes to become a normal weight, than someone who is level 1 overweight, for instance. Plausibility also increases as the original class moves further from the target class, indicating that it is easier to generate a more realistic counterfactual from a query instance that is closer to begin with. Sparsity, diversity and validity do not show any consistent pattern as the class changes. This indicates that these values are not significantly impacted by the original class.

\subsection{Feature Attribution}
\begin{figure*}[ht!]
    \centering
    \begin{subfigure}{0.3\textwidth}\label{attra}
    \includegraphics[width=\linewidth]{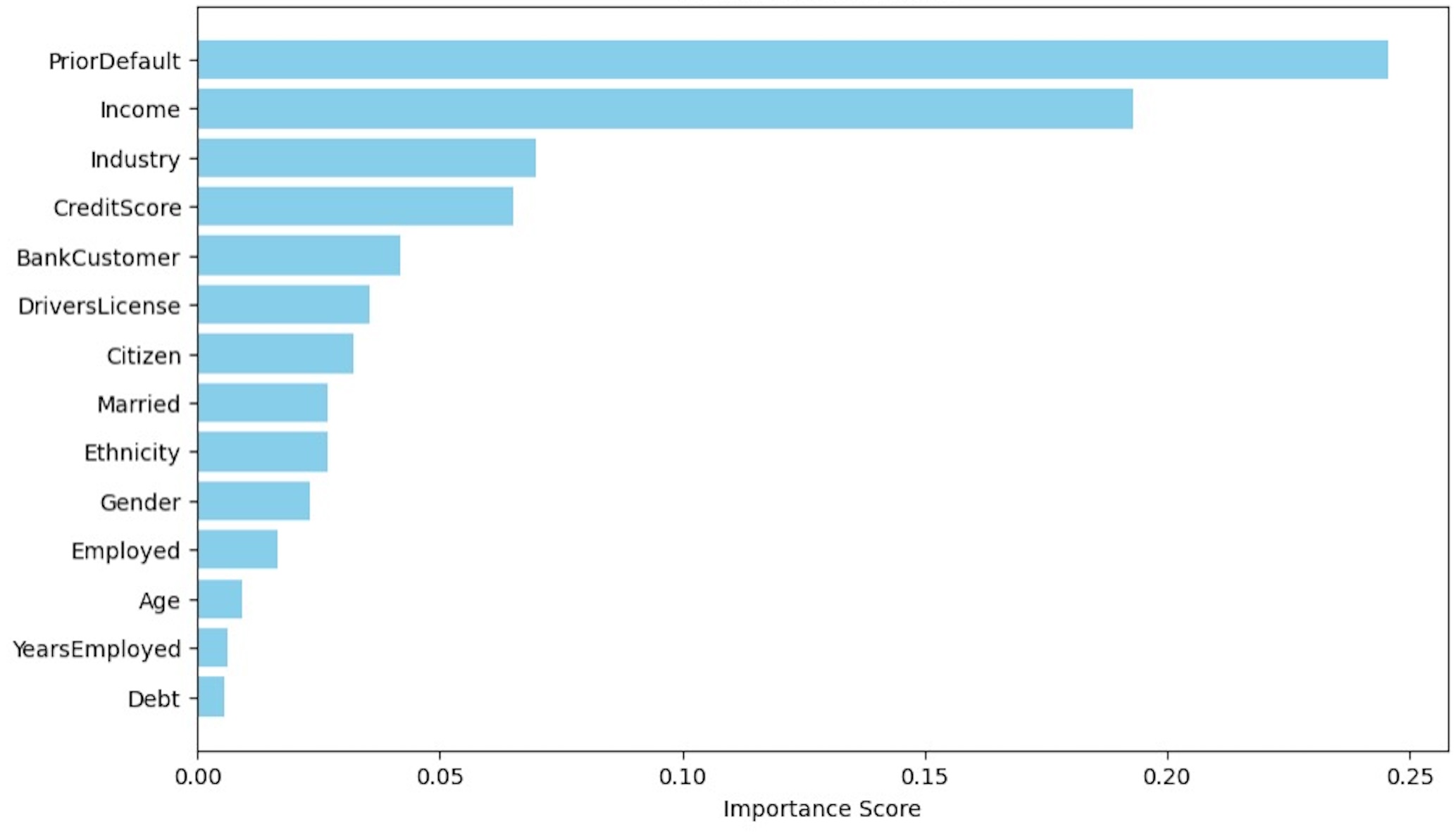}
        \caption{}
    \end{subfigure}
    \begin{subfigure}{0.3\textwidth}\label{attrb}
    \includegraphics[width=\linewidth]{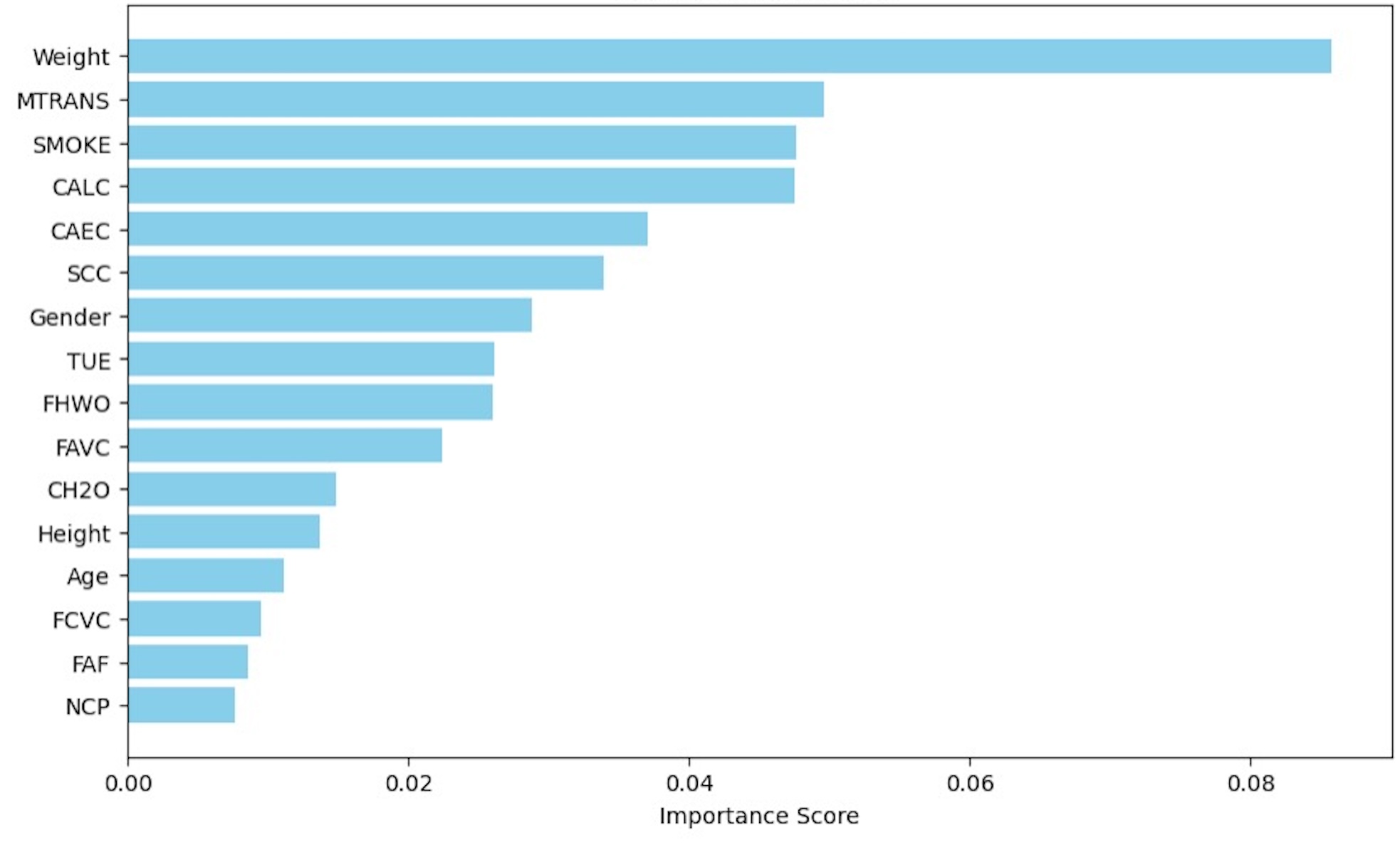}
        \caption{}
    \end{subfigure}
    \begin{subfigure}{0.3\textwidth}\label{attrc}
    \includegraphics[width=\linewidth]{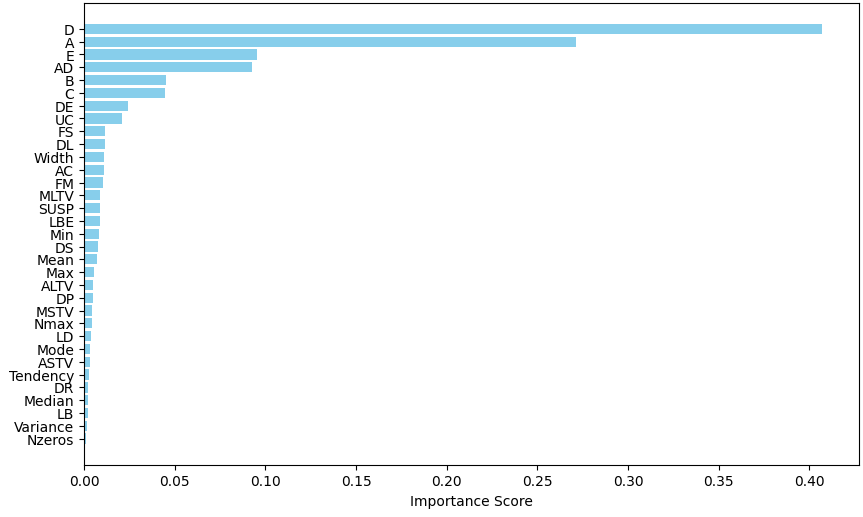}
        \caption{}
    \end{subfigure}
    \caption{FA computed with GradCFA for (a) credit approval, (b) obesity levels and (c) fetal health. The results highlight the critical role of prior default and income in credit approval, weight in obesity classification, and active vigilance and calm sleep in fetal health, while other features like age, years employed, dietary habits, and histogram statistics are less influential in generating plausible and feasible counterfactuals.}
    \label{fig:attr}
\end{figure*}

The FA element of GradCFA informs on the important features for generating feasible, plausible, and diverse counterfactuals, giving deeper insight into model interpretation. An example of the FA computed for each dataset is shown in Fig. \ref{fig:attr}. In credit approval, prior default and income have the highest attribution scores, indicating their crucial role in generating counterfactuals. Conversely, age, years employed, and debt have much lower scores, showing they are less influential. For obesity levels, weight is the most critical feature, as expected. In contrast, age, vegetable consumption frequency (FCVC), physical activity (FAF), and main meals (NCP) have the lowest attribution scores, reflecting their minor role in obesity classification. Lastly, for fetal health, active vigilance (D), and calm sleep (A) have the highest importance scores, while statistics of the histograms such as median, variance and number of zeros (Nzeros) have significantly lower importance.

For comparison, we also compute the feature attribution using state-of-the-art method Shapley Additive Values (SHAP) \cite{lundberg2017a}. In all three cases, the feature deemed most important by SHAP is consistent with the most important feature computed by GradCFA. There is more variance in the less important features, however the scale of the attribution scores is very small, so these differences are not of particular significance. 

\begin{figure*}[ht!]
    \centering
    \begin{subfigure}{0.3\textwidth}\label{shapa}
    \includegraphics[width=\linewidth]{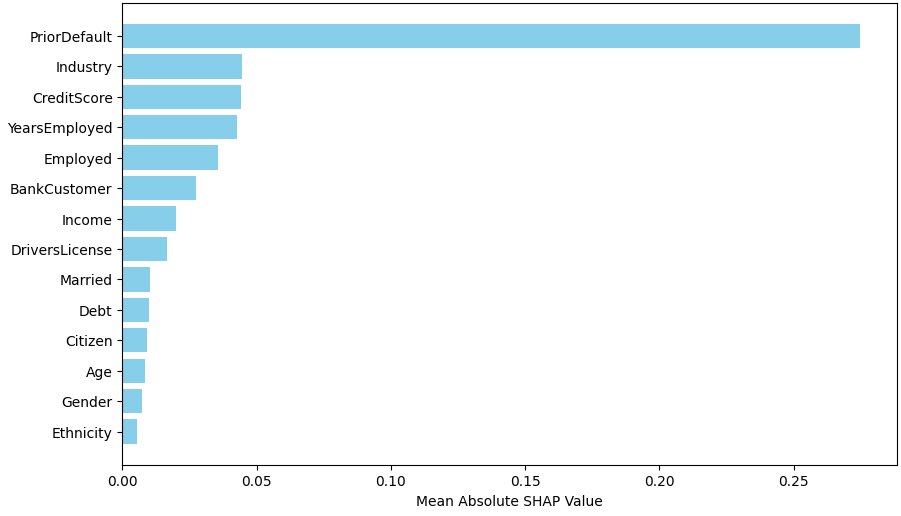}
        \caption{}
    \end{subfigure}
    \begin{subfigure}{0.3\textwidth}\label{shapb}
    \includegraphics[width=\linewidth]{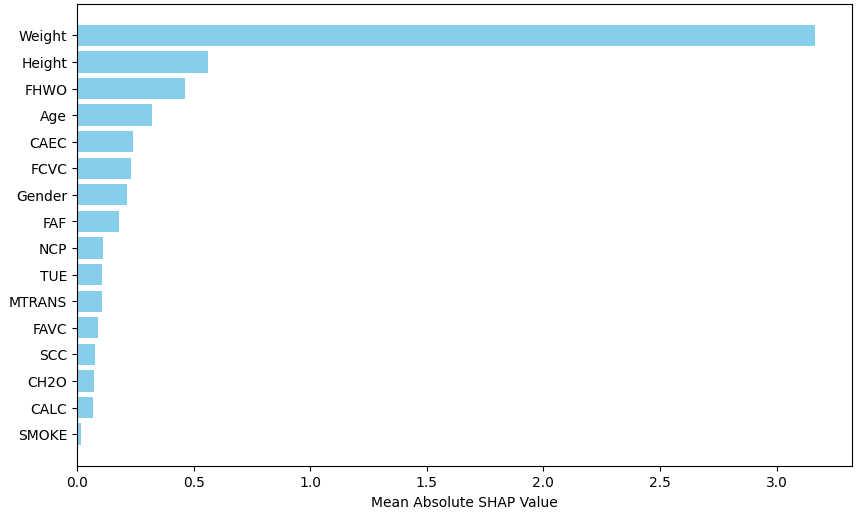}
        \caption{}
    \end{subfigure}
    \begin{subfigure}{0.3\textwidth}\label{shapc}
    \includegraphics[width=\linewidth]{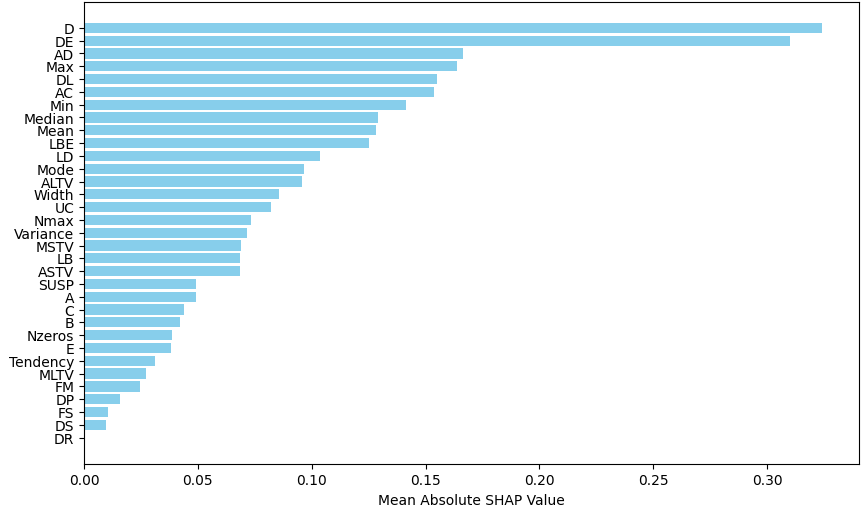}
        \caption{}
    \end{subfigure}
    \caption{FA computed with SHAP for (a) credit approval, (b) obesity levels and (c) fetal health.}
    \label{fig:shap}
\end{figure*}

To assess the FA, we fix each feature and evaluate its impact on counterfactual generation in terms of validity, proximity, sparsity, plausibility, and diversity. Results are presented in Table \ref{tab:fac} for credit approval, Table \ref{tab:fao} for obesity levels, and Table \ref{tab:faf} for fetal health. We exclude the fixed feature from the metric computation to avoid bias. In credit approval, fixing prior default prevents counterfactual generation within the maximum iterations, highlighting its critical role. Removing less influential features, impacts counterfactual quality less significantly. For instance, fixing income (the second most influential feature) results in very low validity (0.54) and weaker metric scores, showing that counterfactuals are less valid and less aligned with desired qualities. This is reflected in a significantly higher plausibility score, indicating the importance of altering income for realistic counterfactuals. Conversely, fixing debt (the least influential feature) has minimal impact, leading to high validity and improved metrics, similar to when all features can vary. For the obesity levels data, fixing the most influential feature renders counterfactual generation impossible, with a validity score of only 0.18, illustrating the complexity of multi-class classification. Fixing the second-most influential feature, mode of transportation (MTRANS), results in weaker metrics and lower validity in counterfactuals, indicating reduced quality. In contrast, fixing less influential features like NCP and FAF yields metrics similar to when all features vary and maintains higher validity, showing minimal impact on counterfactual generation. For the fetal health dataset the top two features have a much higher attribution score than in the other two datasets, and in both cases fixing them prevents successful counterfactual generation. In general it can be observed that as the importance lessens, the counterfactual qualities improve, suggesting that it becomes less challenging to generate actionable counterfactual examples. 

\begin{table}[!ht]
    \centering
    \caption{Analysis of the FA with the Credit Approval Data.}
    \begin{tabular}{cccccc}
    \hline
    Feature & Val. & Prox. & Spars. & Plaus. & Div.\\ Fixed & & & & & \\
    \hline
       PriorDefault & 0.37 & - & - & - & - \\
       Income & 0.54 & 0.26 & 0.24 & 0.87 & 0.82 \\
       Industry & 0.62 & 0.25 & 0.22 & 0.61 & 0.86 \\
       CreditScore & 0.65 & 0.24 & 0.24 & 0.50 & 0.88 \\
       BankCustomer & 0.67 & 0.23 & 0.23 & 0.46 & 0.88 \\
       DriversLicense & 0.68 & 0.22 & 0.24 & 0.45 & 0.91 \\
       Citizen & 0.70 & 0.22 & 0.21 & 0.43 & 0.90 \\
       Married & 0.71 & 0.21 & 0.22 & 0.44 & 0.91 \\
       Ethnicity & 0.73 & 0.21 & 0.21 & 0.42 & 0.93 \\
       Gender & 0.74 & 0.21 & 0.23 & 0.40 & 0.92 \\
       Employed & 0.87 & 0.72 & 0.20 & 0.09 & 0.92 \\
       Age & 0.76 & 0.21 & 0.20 & 0.39 & 0.93 \\
       YearsEmployed & 0.78 & 0.22 & 0.21 & 0.40 & 0.95 \\
       Debt & 0.77 & 0.20 & 0.22 & 0.42 & 0.94 \\
    \hline
    \end{tabular}
    \label{tab:fac}
\end{table}

\begin{table}[!ht]
    \centering
    \caption{Analysis of the FA with the Obesity Levels Data.}
    \begin{tabular}{cccccc}
    \hline
    Feature & Val. & Prox. & Spars. & Plaus. & Div.\\ Fixed & & & & & \\
    \hline
       Weight & 0.18 & - & - & - & - \\
       MTRANS & 0.52 & 0.33 & 0.42 & 0.62 & 0.71 \\
       SMOKE & 0.57 & 0.29 & 0.37 & 0.67 & 0.75 \\
       CALC & 0.59 & 0.28 & 0.35 & 0.61 & 0.76 \\
       CAEC & 0.61 & 0.28 & 0.30 & 0.55 & 0.78 \\
       SCC & 0.58 & 0.27 & 0.31 & 0.58 & 0.81 \\
       Gender & 0.62 & 0.26 & 0.30 & 0.59 & 0.80 \\
       TUE & 0.65 & 0.25 & 0.33 & 0.56 & 0.79 \\
       FHWO & 0.66 & 0.25 & 0.31 & 0.54 & 0.82 \\
       FAVC & 0.67 & 0.24 & 0.30 & 0.53 & 0.81 \\
       \hline
       CH2O & 0.69 & 0.23 & 0.28 & 0.58 & 0.84 \\
       Height & 0.70 & 0.22 & 0.27 & 0.52 & 0.85 \\
       Age & 0.71 & 0.23 & 0.26 & 0.53 & 0.83 \\
       FCVC & 0.72 & 0.24 & 0.25 & 0.49 & 0.84 \\
       FAF & 0.74 & 0.23 & 0.24 & 0.46 & 0.83 \\
       NCP & 0.75 & 0.22 & 0.24 & 0.45 & 0.85 \\
    \hline
    \end{tabular}
    \label{tab:fao}
\end{table}

\begin{table}[!ht]
    \centering
    \caption{Analysis of the FA with the Fetal Health Data.}
    \begin{tabular}{cccccc}    
    \hline
    Feature & Val. & Prox. & Spars. & Plaus. & Div.\\ Fixed & & & & & \\
    \hline     
         D & 0.24 & - & - & - & - \\
         A & 0.45 & - & - & - & - \\
         AD & 0.51 & 0.48 & 0.49 & 0.77 & 0.79 \\
         B & 0.53 & 0.47 & 0.47 & 0.71 & 0.77 \\
         C & 0.56 & 0.47 & 0.46 & 0.69 & 0.80 \\
         DE & 0.57 & 0.47 & 0.47 & 0.67 & 0.81 \\
         UC & 0.58 & 0.46 & 0.46 & 0.66 & 0.83 \\
         FS & 0.60 & 0.45 & 0.45 & 0.65 & 0.83 \\
         DL & 0.61 & 0.46 & 0.46 & 0.67 & 0.82 \\
         Width & 0.60 & 0.46 & 0.44 & 0.64 & 0.82 \\
         AC & 0.62 & 0.43 & 0.45 & 0.63 & 0.84 \\
         \hline
         FM & 0.63 & 0.44 & 0.43 & 0.62 & 0.85 \\
         MLTV & 0.64 & 0.44 & 0.45 & 0.61 & 0.85 \\
         SUSP & 0.65 & 0.43 & 0.43 & 0.62 & 0.86 \\
         LBE & 0.66 & 0.41 & 0.43 & 0.59 & 0.87 \\
         Min & 0.66 & 0.43 & 0.42 & 0.60 & 0.87 \\
         DS & 0.67 & 0.42 & 0.42 & 0.58 & 0.88 \\
         Mean & 0.67 & 0.43 & 0.44 & 0.57 & 0.88 \\
         Max & 0.68 & 0.41 & 0.44 & 0.55 & 0.88 \\
         ALTV & 0.69 & 0.42 & 0.42 & 0.54 & 0.89 \\
         DP & 0.68 & 0.42 & 0.41 & 0.53 & 0.89 \\
         MSTV & 0.69 & 0.41 & 0.41 & 0.52 & 0.89 \\
         Nmax & 0.69 & 0.41 & 0.40 & 0.51 & 0.89 \\
         LD & 0.70 & 0.41 & 0.43 & 0.50 & 0.90 \\
         Mode & 0.70 & 0.40 & 0.41 & 0.49 & 0.90 \\
         ASTV & 0.71 & 0.40 & 0.40 & 0.48 & 0.90 \\
         Tendency & 0.71 & 0.41 & 0.40 & 0.47 & 0.90 \\
         DR & 0.72 & 0.40 & 0.40 & 0.47 & 0.90 \\
         Median & 0.72 & 0.40 & 0.49 & 0.48 & 0.91 \\
         LB & 0.73 & 0.30 & 0.39 & 0.44 & 0.91 \\
         Variance & 0.73 & 0.39 & 0.40 & 0.44 & 0.91 \\
         Nzeros & 0.73 & 0.36 & 0.38 & 0.42 & 0.93 \\
    \hline
    \end{tabular}
    \label{tab:faf}
\end{table}

\section{Discussion}
Across all experiments, GradCFA consistently generates balanced counterfactuals in terms of proximity, sparsity, plausibility, and diversity, while maintaining high classification confidence. Our results show that GradCFA’s loss function and optimization strategy outperform other configurations across binary and multi-class problems, emphasizing the importance of each element in achieving this balance. This quality balance is crucial for practical utility, ensuring counterfactuals are both feasible and provide a diverse range of realistic options.

A major contribution of GradCFA is its FA mechanism, which identifies influential features crucial for generating feasible, plausible, and diverse counterfactuals. FA analysis reveals that certain features—such as prior default and income in credit approval, weight in obesity levels, and vigilance measures in fetal health—are significantly more impactful. The validation of these FA results by setting each feature to immutable confirms these findings, as fixing high-impact features typically prevents or degrades counterfactual generation. For instance, in credit approval, setting prior default to immutable halts generation within maximum iterations, underscoring its significance. Similarly, in obesity levels, fixing weight reduces confidence and validity, and in fetal health, the top two features prevent generation when fixed. These validations underscore FA’s utility in identifying critical features, offering stakeholders targeted insights on the most impactful variables for achieving desired outcomes.

GradCFA provides extensive user control, making it not only robust in terms of metrics but also adaptable for real-world applications. Through our ablation study, we show that the complete GradCFA loss function achieves an effective balance across metrics, but users can tailor the weights of each term based on specific needs. For example, prioritizing diversity allows users to explore a broader range of options, while focusing on proximity and sparsity suits users with limited resources who require minimal feature changes. GradCFA also features threshold and scale factor hyperparameters, enabling users to specify a threshold for acceptable loss and apply perturbations to escape local optima when needed. GradCFA’s support for user-defined constraints, such as setting ranges and directional limits for certain features, ensures that only realistic, application-specific counterfactuals are generated.

The analysis of example counterfactuals demonstrates GradCFA’s utility in generating actionable insights. In the credit approval dataset, counterfactuals suggest that increasing employment status and years employed are critical for approval, aligning with expectations. However, certain unintuitive findings, such as increased debt or default status in some counterfactuals, highlight potential areas for model refinement or complex feature dependencies. These findings illustrate GradCFA’s dual value for stakeholders and data scientists, offering actionable insights and aiding model improvement.

GradCFA’s performance on multi-class problems also reveals insights into how different problem characteristics affect CFX generation. In binary classification cases like credit approval, achieving sparse and diverse counterfactuals is relatively straightforward due to a single decision boundary. In multi-class problems, such as obesity levels and fetal health, complexity increases due to multiple decision boundaries and intricate feature interactions. This is reflected in generally higher proximity and sparsity values in multi-class cases, especially for the fetal health dataset, where the high proportion of continuous features makes slight changes more likely, leading to increased proximity and sparsity. Nonetheless, the greater feature count does not negatively impact counterfactual quality, as comparable validity, plausibility, and diversity scores are observed across datasets. This consistency suggests GradCFA’s robustness, even as feature counts increase, reinforcing its suitability for high-dimensional applications.

\section{Conclusion}
In this paper, we introduced GradCFA, a novel CFX algorithm that combines counterfactual generation with FA to enhance model interpretability. Through experiments across binary and multi-class classification tasks, we showed that GradCFA's loss function and optimization strategy outperform existing methods, offering actionable insights for stakeholders and valuable information for data scientists. The FA mechanism highlights influential features for counterfactual generation, providing a deeper understanding of feature impact on model decisions and enabling targeted model improvements.

A key strength of GradCFA is its flexibility, allowing customization of hyperparameters to suit specific needs and user-defined constraints, to enhance the feasibility and relevance of the generated counterfactuals. However, we recognize that selecting appropriate hyperparameters may pose challenges for non-expert users. Future work could address this by developing a user-centered interface to assist in hyperparameter selection, potentially incorporating interactive visualizations or automated tuning mechanisms. Such an interface would align GradCFA with broader user-centered design principles, further supporting its adoption in real-world applications.

While GradCFA’s current gradient-based optimization is effective, it can be computationally intensive for large datasets or complex models. Future work could explore more efficient optimization strategies to improve scalability. Additionally, while GradCFA generates statistically plausible counterfactuals, it does not yet fully account for causal relationships.

We see significant potential in integrating Graph Neural Networks (GNNs) to enhance causal awareness in GradCFA. GNNs, as demonstrated in recent work \cite{metsch2024clarus}, can model complex dependencies between features and support interactive counterfactual queries, making them promising for explainable AI. Incorporating GNN architectures could improve GradCFA’s capacity to generate nuanced counterfactuals that account for underlying causal structures and allow direct causal reasoning within the model.

\bibliographystyle{ieeetr}
\bibliography{bibliography}

\end{document}